\documentclass[runningheads]{llncs}
\usepackage{graphicx}
\usepackage{xcolor}
\usepackage{csquotes}
\usepackage{mdframed}
\newcommand{\reasonx}[1]{\normalfont {\scshape reasonx}}
\usepackage{dsfont}

\usepackage{adjustbox}
\usepackage{hyperref}

\makeatletter
\renewcommand{\paragraph}{%
  \@startsection{paragraph}{4}%
  {\z@}{1.25ex \@plus 1ex \@minus .2ex}{-0.3em}%
  {\normalfont\normalsize\bfseries}%
}
\makeatother

\begin{document}
\title{Declarative Reasoning on Explanations Using Constraint Logic Programming 
}
\titlerunning{Declarative Reasoning on Explanations Using CLP}
%
\author{Laura State\inst{1,2}\orcidID{0000-0001-8084-5297} \and
Salvatore Ruggieri\inst{1}\orcidID{0000-0002-1917-6087} \and
Franco Turini\inst{1}\orcidID{0000-0001-6789-5476}}
%

%
%
\institute{University of Pisa, Pisa, Italy \and
Scuola Normale Superiore, Pisa, Italy
}

\maketitle              
\begin{abstract}

Explaining opaque Machine Learning (ML) models is an increasingly relevant problem.
Current explanation in AI (XAI) methods suffer several shortcomings, among others an insufficient incorporation of background knowledge, and a lack of abstraction and interactivity with the user.
We propose \reasonx{}, an explanation method based on Constraint Logic Programming (CLP).
\reasonx{} can provide declarative, interactive explanations for decision trees, which can be the ML models under analysis or global/local surrogate models of any black-box model.
Users can express background or common sense knowledge using linear constraints and MILP optimization over features of factual and contrastive instances, and interact with the answer constraints at different levels of abstraction through constraint projection. We present here the architecture of \reasonx{}, which consists of a Python layer, closer to the user, and a CLP layer. \reasonx{}'s core execution engine is a Prolog meta-program with declarative semantics in terms of logic theories.
\end{abstract}

\section{Introduction}

Artificial Intelligence (AI) systems are increasingly being adopted for taking critical decisions impacting society, such as loan concession in bank systems. 
The acceptance and trust of applications based on AI is hampered by the opaqueness and complexity of the Machine Learning (ML) models adopted, possibly resulting in biased or socially discriminatory decision-making~\cite{DBLP:journals/widm/NtoutsiFGINVRTP20}.

For these reasons, there has recently been a flourishing of proposals for explaining the decision rationale of ML models~\cite{DBLP:journals/csur/GuidottiMRTGP19,DBLP:journals/ai/Miller19,DBLP:journals/air/MinhWLN22,Molnar2020}, coined eXplanation in AI (XAI) methods.
These approaches lack sufficient abstraction for reasoning over the decision rationale of the ML model. By reasoning, we mean the possibility for the user to define any number of conditions over factual and contrastive instances, which would codify both background knowledge and what-if analyses, and then looking at answers at the symbolic and intensional level. 

To close this gap, we present \reasonx{} (\textit{reason to explain}), an explanation tool built in two layers. The first is in Python, closer to users, where decision tree (DT) models and user queries are parsed and translated. 
The DT can be the ML model itself, or a surrogate of other ML models at global/local level. 
The second is in Constraint Logic Programming (CLP), where embedding of DTs and background knowledge are reasoned about, using a Prolog meta-program.

We display an exemplary dialogue between a fictional user and \reasonx{} below.
It is situated in the context of a credit application scenario, i.e. the user is a person whose credit application has been rejected by an automated decision-making system.
Please note that while the information content is exactly what \reasonx{} can provide, we enhanced the dialogue by translating the interaction into natural language, to mimic better a realistic interaction.
\begin{mdframed}
\small
\texttt{USER}: 
Can I see the rule that led to the denial of my credit application? \\
\texttt{REASONX}: Your credit application was rejected, because your income is lower than 60,000 EUR/year, and you still have to pay back the lease of your car. \\
\texttt{USER}: Ok. Can you present me two options that will lead to a change of the decision outcome? Please take into consideration that I need a credit of at least 10,000 EUR. 
{I would like to see options that require as little change as necessary.} \\
\texttt{REASONX}: You have the following two options: 
You pay back the lease on the car, or you increase your age by 10 years (from 35 to 45 years). \\
\texttt{USER}: The second option presented is a bit strange. I am wondering whether this is salient in the model. 
Can I please see the options to obtain credit for an individual with the same properties as me, for a credit of at least 10,000 EUR, but with the feature age at 35 years or less (i.e. young applicant), instead of fixed? \\
\texttt{REASONX}: For the given profile, the credit is always rejected. \\
\texttt{USER}:
Given this profile, how can the decision reversed, under as little changes as possible?\\
\texttt{REASONX}: Credit can be obtained, if the feature age is set to higher than 35 years. \\
\texttt{USER}: This interesting and worth investigating further. There could be bias w.r.t. the age of the person that applies for credit.
\end{mdframed}
Adding background knowledge to explanations has the potential to significantly improve their quality~\cite{DBLP:journals/corr/abs-2105-10172,DBLP:conf/iclp/State21}. 
Ignoring it can lead to explanations that disregard the needs of the user, or do not fit the reality of our world - depending on its purpose.
An example is the minimum credit amount (\enquote{a credit of at least 10,000 EUR}).
Further, interactivity arises naturally in \reasonx{}: the user can flexibly query it, choosing queries that best fit to her questions, e.g., by adding constraints, and thereby building an own, personalized explanation.

Here, we focus on the CLP layer of \reasonx{}. 
The Python layer and case studies at the user level are thoroughly presented in a companion paper~\cite{xaiworld}.

The paper is structured as follows. In Sec. \ref{sec:background}, we discuss background and related work. Sec. \ref{sec:clp} describes the syntax, semantics, and meta-programming features of CLP that \reasonx{} builds on. The architecture of \reasonx{} is described in Sec. \ref{sec:reasonx}. We summarize contributions and future work in Sec. \ref{sec:conclusion}. 

\section{Background and Related Work}
\label{sec:background}\label{sec:related_work}

\paragraph{Logic and Knowledge in XAI.} 

Several XAI approaches have used (propositional) logic rules as forms of model-agnostic explanations both locally \cite{DBLP:journals/expert/GuidottiMGPRT19,DBLP:conf/aaai/Ribeiro0G18,DBLP:journals/tvcg/MingQB19} and globally \cite{DBLP:journals/ai/SetzuGMTPG21}. 
Such approaches, however, do not allow for reasoning over produced explanations. 
Surveys on work at the intersection between symbolic and sub-symbolic methods (incl. argumentation and abduction) are~\cite{DBLP:journals/ia/CalegariCO20,DBLP:journals/corr/abs-2010-12896,DBLP:journals/frai/DietzKM22}.

\paragraph{Contrastive Explanations.} 

Contrastive explanations\footnote{To avoid confusion with the concept of counterfactuals as understood in the statistical causality literature, and following~\cite{DBLP:journals/ai/Miller19}, we use the term contrastive explanations.} (CEs), i.e., instances similar to those to explain but with different labels assigned by the black-box (BB) classifier, are a key element in causal approaches to interpretability~\cite{DBLP:journals/inffus/ChouMBOJ22,DBLP:journals/access/StepinACP21}. 
~\cite{DBLP:journals/corr/abs-1711-00399} introduces contrastive explanations to the field of XAI, with several extensions~\cite{DBLP:conf/ijcai/KanamoriTKA20,DBLP:conf/fat/Russell19}.
Moreover, while \cite{DBLP:conf/ijcai/Byrne19} argues in favor of CEs from a psychological point of view, ~\cite{DBLP:journals/ai/Miller19,DBLP:conf/fat/MittelstadtRW19} make clear that explanations in a contrastive form are highly desirable for (lay) end-users.

\paragraph{Interactivity.} 

Interactivity aligns closely with our working definition of an explanation: \enquote{[...] an interaction, or an exchange of information}, where it crucially matters to \textit{whom} the explanation is given, and for \textit{what} purpose~\cite{DBLP:conf/iclp/State21}.
\cite{DBLP:journals/ki/SokolF20} convincingly arguments for interactivity by presenting the glass-box tool~\cite{DBLP:conf/ijcai/SokolF18a}.
~\cite{DBLP:journals/corr/abs-2202-01875} confirms the relevance of interactivity via an interview study with practitioners.

\paragraph{Explanations and Decision Trees.}

Closely linked work is presented by a series of papers of Sokol et al., introducing explanations for DTs~\cite{sokol2021intelligible}, generalizing it to local surrogate models~\cite{DBLP:journals/corr/abs-2005-01427}, and exploiting interactivity~\cite{DBLP:conf/ijcai/SokolF18a}.
Again, the main difference to our work is our reliance on CLP, and thus reasoning capabilities.
Another related work is \cite{DBLP:journals/corr/abs-2011-07423}, providing CEs via (actual) causality. 

\paragraph{Embedding Decision Trees into Constraints.}

In this paper, we assume that the DT is already available. We reason over the DT by encoding it as a set of linear constraints. This problem, known as embedding \cite{DBLP:conf/cpaior/BonfiettiLM15}, requires to satisfy $c(x, y) \Leftrightarrow f(x) = y$,
where $f(x)$ is the class as predicted by the DT, $x$ the input vector consisting of discrete and/or continuous variables, and $c$ is a constraint of some type. We adopt a rule-based encoding, which takes space in $O(N \log{N})$ where $N$ is the number of nodes in the DT. Other encodings, such as Table and MDD \cite{DBLP:conf/cpaior/BonfiettiLM15}, require discretization of continuous features, thus losing the power of reasoning over linear constraints over reals.

\section{Preliminaries: Constraint Logic Programming}\label{sec:clp}

Logic programming (LP) is a declarative approach to problem-solving based on logic rules in the form of Horn clauses~\cite{Apt1997}. It supports reasoning under various settings, e.g., deductive, inductive, abductive, and meta-reasoning~\cite{DBLP:journals/jair/CropperD22,10.5555/773294}. 
Starting with Prolog~\cite{DBLP:books/daglib/0076175}, LP has been extended in several directions, as per expressivity and efficiency ~\cite{DBLP:journals/tplp/KornerLBCDHMWDA22}. 
Constraint logic programming (CLP) augments logic programming with the ability to solve constrained problems~\cite{DBLP:journals/toplas/JaffarMSY92}. 
The CLP scheme defines a family of languages, CLP($\cal C)$, that is parametric in the constraint
domain $\cal C$. We are interested in CLP($\cal R)$, namely the constraint domain over the
reals. We use the SWI Prolog system~\cite{DBLP:journals/tplp/WielemakerSTL12} implementation. 

We rely on meta-programming, a powerful technique that allows a LP to manipulate programs encoded as terms. This is extended in CLP by encoding constraints as terms.  

Further, CLP($\cal R)$ offers mixed integer linear programming (MILP) optimization functionalities ~\cite{DBLP:phd/ndltd/Magatao10}. Common predicates include the calculation of the supremum and the infimum of an expression w.r.t. the solutions of the constraint store.
Complex constraint meta-reasoning procedures are based on such predicates, some examples are~\cite{DBLP:journals/tplp/BenoyKM05,DBLP:conf/ecai/Ruggieri12}.

\section{Explaining via Reasoning: \reasonx{}}
\label{sec:reasonx}

\reasonx{} consists of two layers.
The top layer in Python is designed for integration with the \texttt{pandas} and \texttt{scikit-learn} standard libraries for data storage and model construction. Meta-data, models, and user constraints specified at this level are parsed and transformed into Prolog facts.
The bottom layer is in CLP($\cal R$) and it is written in SWI Prolog~\cite{DBLP:journals/tplp/WielemakerSTL12}. 

\reasonx{} relies on a DT, the \textit{base model}. Such a tree can be: (a) the model to be explained/reasoned about\footnote{While DTs are generally thought interpretable, it depends on their size/depth. Large DTs are hard to reason about, especially in a contrastive explanation scenario.}; (b) a global surrogate of an opaque ML model; (c) a local surrogate trained to mimic a BB model in the neighborhood of the (local) instance to explain. 
In cases (b) and (c), the surrogate model is assumed to have good fidelity in reproducing the decisions of the black-box. This is reasonable for local models, i.e., in case (c). Learning the tree over a local neighborhood has been a common strategy in perturbation-based XAI methods such as LIME~\cite{DBLP:conf/kdd/Ribeiro0G16}.
Following, we present an excerpt of the initialization code: 
\\[1ex]
\texttt{\small > r = reasonx.ReasonX($\ldots$)}\\
\texttt{\small > r.model(clf)}\\[1ex]
where the meta-data about the features are passed to the object \texttt{\small r} during its creation, and the DT \texttt{\small clf} is passed over. There can be more than one base model to account for different ML models, e.g., Neural Networks vs ensembles.
The user can declare and reason about one or more instances, factual or contrastive, by specifying their class value. Each instance refers to a specific base model. The instance does not need to be fully specified, as in existing XAI methods. For example, an instance \texttt{F} can be declared with only the following characteristics:   
\\[1ex]
\texttt{\small > r.instance(`F', label=1)}\\
\texttt{\small > r.constraint(`F.age = 30, F.capitalloss >= 1000')}\\[1ex]
to intensionally denote a persons with age of 30 and capital loss of at least $1,000$.
Background knowledge can be expressed through linear constraints over features of instances. E.g., by declaring another instance \texttt{CE} classified differently by the base model (the contrastive instance), the following constraints require that the contrastive instance must not be younger, and has a larger capital loss:\\[1ex]
\texttt{\small > r.instance(`CE', label=0)}\\
\texttt{\small > r.constraint(`F.age <= CE.age, CE.capitalloss >= F.capitalloss + 500')}\\[1ex]
The output of \reasonx{} consists of constraints for which the declared instances are classified as expected by the DT(s) and such that user and implicit constraints on feature data types are satisfied. The output can be projected on only some of the instances or of the features:\\[1ex]
\texttt{\small > r.solveopt(project=[`CE'])}\\
\texttt{\small > ---}\\
\texttt{\small > Answer: 30 <= CE.age, F.capitalloss >= 1500, CE.hoursperweek >= 40.0}\\[1ex]
where \texttt{\small 30 <= CE.age, F.capitalloss >= 1500
} are entailed by the constraints and \texttt{\small CE.hoursperweek >= 40.0} is due to conditions in the DT.  Moreover, the user can specify a distance function for the minimization problem to derive the closest contrastive example, e.g., as in \texttt{\small  solveopt(minimize=`l1norm(F, CE)')}.

\subsection{Embeddings into CLP} \label{sec:bblock}

We are agnostic about the learning algorithm that produces the base model(s). 
Features can be nominal, ordinal, or continuous. Ordinal features are coded as consecutive integer values (some preprocessing is offered in \reasonx{}).
Nominal features can be one-hot encoded or not.
When embedding the DT into CLP, we force one-hot encoding of nominal features anyway, and silently decode back when returning the answer constraints to the user. A nominal feature $x_i$ is one-hot encoded into $x_i^{v_1}, \ldots, x_i^{v_k}$ with $v_1, \ldots, v_k$ being the distinct values in the domain of $x_i$.
We assume that the split conditions from a parent node to a child node are of the form $\mathbf{a}^T \mathbf{x} \simeq b$, where $\mathbf{x}$ is the vector of features $x_i$'s. The following common split conditions are covered by such an assumption:
\begin{itemize}
    \item axis-parallel splits for continuous and ordinal features, i.e., $x_i \leq b$ or $x_i > b$;
    \item linear splits for continuous features: ${\bf a}^T{\bf x} \leq b$ or ${\bf a}^T{\bf x} > b$;
    \item (in)equality splits for nominal features: $x_i = v$ or $x_i \neq v$; in terms of one-hot encoding, they respectively translate into  $x_i^v = 1$ or $x_i^v = 0$.
\end{itemize}
Axis parallel and equality splits are used in CART \cite{DBLP:books/wa/BreimanFOS84} and C4.5 \cite{DBLP:books/mk/Quinlan93}. Linear splits are used in oblique \cite{DBLP:journals/jair/MurthyKS94} and optimal decision trees \cite{DBLP:journals/ml/BertsimasD17}. Linear model trees combine axis parallel splits at nodes and linear splits at leaves \cite{DBLP:journals/ml/FrankWIHW98}.

\paragraph{Embedding base model(s) into Prolog Facts.}
\label{sec:translation_model_to_facts}

Each path (root to the leaf in the DT), is translated into a fact, a conjunction of linear split conditions:
\begin{center}\small
\texttt{path($m$,[$\mathbf{x}$], [$\mathbf{a}_1^T \mathbf{x} \simeq b_1$, $\ldots$, $\mathbf{a}_k^T \mathbf{x} \simeq b_k$], $c$, $p$).}
\end{center}
where $m$ is an id of the decision tree, \texttt{[$\mathbf{x}$]} a list of (Prolog) variables representing the features, $c$ the class predicted at the leaf, $p$ the confidence of the prediction, and \texttt{[$\mathbf{a}_1^T \mathbf{x} \simeq b_1$, $\ldots$, $\mathbf{a}_k^T \mathbf{x} \simeq b_k$]} the list of the $k$ split conditions.

\paragraph{Encoding instances.}
Each instance is represented by a list of Prolog variables.
The mapping between names and variables is positional, and decoding is stored in a predicate \texttt{feature($i$, $\mathit{varname}$)} where $i$ is a natural number and $\mathit{varname}$ a constant symbol, e.g., \texttt{vAge}. All instances are collectively represented by a list of lists of variables $\mathit{vars}$. 
Further, \reasonx{} is defining a utility predicate \texttt{data\_instance} with instance's meta-data.

\paragraph{Encoding implicit constraints ($\Psi$).}
Constraints on the features $\mathbf{x}$ of each instance derive from their data types. We call them ``implicit'' because the system can generate them from meta-data:
\begin{itemize}
    \item for continuous features: $x_i \in {\cal R}$;
    \item for ordinal features: $x_i \in {\cal Z}$ and $m_i \leq x_i \leq M_i$ where $dom(x_i) = \
    \{m_i, \ldots, M_i\}$;
    \item for one-hot encoded nominal features: $x^{v_1}_i, \ldots, x^{v_k}_i \in {\cal Z}$ and $\wedge_{j=1}^k 0 \leq x^{v_j}_i \leq 1$ and $\sum_{j=1}^k x^{v_j}_i = 1$;
\end{itemize}
Constraints for ordinal and nominal features are computed by the Prolog predicates \texttt{ord\_constraints($\mathit{vars}$, COrd)} and \texttt{cat\_constraints($\mathit{vars}$, CCat)} respectively.
We denote by $\Psi$ the conjunction of all implicit constraints.

\paragraph{Encoding user constraints ($\Phi$).} The following background knowledge, loosely categorized as in~\cite{DBLP:journals/corr/abs-2010-04050}, can be readily expressed in \reasonx{}:
\begin{description}
\item[{\normalfont \textit{Feasibility}}]
Constraints concerning the possibility of feature changes, and how these depend on previous values or (changes of) other features:
\begin{itemize}
    \item \textit{Immutability}: a feature cannot/must not change. 
    \item \textit{Mutable but not actionable}: the change is only a result of changes in features it depends upon. 
    \item \textit{Actionable but constrained}: 
    the feature can be changed only under some condition. 
\end{itemize}
\item[{\normalfont \textit{Consistency}}] Constraints aiming at specific domain values a feature can take. 
\end{description}
Constraints specified in Python are parsed and transformed into a list of CLP constraints. 
An interpreter of expressions is provided which returns a list of linear constraints over variables.
The only non-linear constraint is equality of nominal values and is translated exploiting one-hot-encoding of nominal features.

\paragraph{Encoding distance functions.}

We simplify the optimization proposed in \cite{DBLP:journals/corr/abs-1711-00399} by the assumption that declared instances have a class label\footnote{The split conditions from root to leaf do not necessarily lead to the same class label with 100\% probability. \reasonx{} includes a parameter in the declaration of an instance to require a minimum confidence value of the required class.}. 
The distance function is defined as a linear combination of $L_1$ and $L_{\infty}$ norms for ordinal and continuous features and of a simple matching distance for nominal features:

\begin{eqnarray}\label{eq:dist}\small
\min \hspace{-0.2cm}  \sum_{i\ \textrm{\tiny nominal}} \mathds{1} (x_{cf, i} \ne x_{f, i}) + \beta\hspace{-0.2cm} \sum_{i\ \textrm{\tiny ord., cont.}}  | x_{cf, i} - x_{f, i} | + \gamma \hspace{-0.2cm} \max_{i\ \textrm{\tiny ord., cont.}}  | x_{cf, i} - x_{f, i} |\label{equ:opt2}
\end{eqnarray}
where $\beta$ and $\gamma$ denote parameters. $L_1$ and $L_{\infty}$ norms are calculated over max-min normalized values to account for different units of measures. 
See \cite{DBLP:conf/aistats/KarimiBBV20,DBLP:journals/corr/abs-1711-00399} for a discussion.
To solve the MILP problem, we need to linearize the minimization. This leads to additional constraints and slack variables. 

\subsection{The Core Meta-interpreter of \reasonx{}}\label{sec:meta}

We reason on constraints as theories and design operators for composing such theories. The core engine of \reasonx{} is implemented as a Prolog meta-interpreter of expressions over those operators.

A (logic) theory is a set of formulas, from which one is interested to derive implied formulas, and a logic program is itself a theory \cite{DBLP:conf/iclp/BrogiMPT91}. In our context, a theory consists of a set of linear constraints $\{c_i\}_i$ to be interpreted as the disjunction $\vee_i \ c_i$. Theories are coded in LP by exploiting its non-deterministic computational model, i.e., each $c_i$'s is returned by a clause in the program.
The language of expressions over theories is closed: operators map one or more theories into a theory. 
The following theories are included:
\begin{description}
    \item[\texttt{typec}]  the theory with only the conjunction $\wedge_{c \in \Psi}\ c$ of the implicit constraints;
    \item[\texttt{userc}] the theory with only the conjunction $\wedge_{c \in \Phi}\ c$ of the user constraints;
    \item[\texttt{inst(I)}]  the theory of constraints $\wedge_i\ \mathbf{a}_i^T \mathbf{x} \simeq b_i$ where $\mathbf{x}$ are features of the instance \texttt{I}, and primitive constraints $\mathbf{a}_i^T \mathbf{x} \simeq b_i$ are those in the path of the decision tree \texttt{M} the instance refers to. 
\end{description}

We provide the following operators on theories: the cross-product of constraints of theories, the subset of constraints in a theory that are satisfiable, the projection of constraints in a theory over a set of variables, and the subset of constraints in a theory that minimize a certain (distance) function.

The queries to the CLP layer of \reasonx{} can be answered by a  Prolog query over the predicates \texttt{instvar} (building $vars$), \texttt{proj\_vars} (computing which of those variables are to be projected in the output), and \texttt{solve} (evaluating expressions over the cross-product of \texttt{typec}, \texttt{userc}, and the theories \texttt{inst(I)} for all instances \texttt{I}).

\section{Conclusion}
\label{sec:conclusion}

We presented REASONX, a declarative XAI tool that relies on linear constraint reasoning,  
solving for 
background knowledge, and for interaction with the user at a high abstraction and intensional level.
These features make it a unique tool when compared to instance-level approaches commonly adopted for explaining ML models.
We aim at extending \reasonx{} along three directions: 
\textit{i)} the implementation of additional constraints, possibly with non-linear solvers,
\textit{ii)} an extensive evaluation based on some theoretical measures, as well as through user-studies~\cite{DBLP:journals/corr/abs-2210-11584} and real-world data,
and \textit{iii)} extension to non-structured data, such as images and text, e.g., through the integration of concepts~\cite{DBLP:conf/aiia/DonadelloD20}.

{\small \paragraph{Software.}
\reasonx{} is released open source at \href{https://github.com/lstate/REASONX}{https://github.com/lstate/REASONX}.}

{\small \paragraph{Acknowledgments.}
Work supported by the European Union’s Horizon 2020 research and innovation programme under Marie Sklodowska-Curie Actions for the project NoBIAS 
(g.a. No. 860630), and by the NextGenerationEU program within the PNRR-PE-AI scheme (M4C2, investment 1.3, line on Artificial Intelligence) for the project FAIR (Future Artificial Intelligence Research).
This work reflects only the authors’ views and the European Research Executive Agency (REA) is not responsible for any use that may be made of the information it contains.}

%
%
%
\bibliographystyle{splncs04}
\bibliography{library}
\end{document}